\documentclass[runningheads]{llncs}
\usepackage{graphicx}

\usepackage{tikz}
\usepackage{comment} 
\usepackage{amsmath,amssymb} %
\usepackage{color}

\usepackage{bm}
\usepackage{multirow} %
\usepackage{ctable}   %

\newcolumntype{L}[1]{>{\raggedright\let\newline\\\arraybackslash\hspace{0pt}}m{#1}}
\newcolumntype{C}[1]{>{\centering\let\newline\\\arraybackslash\hspace{0pt}}m{#1}}
\newcolumntype{R}[1]{>{\raggedleft\let\newline\\\arraybackslash\hspace{0pt}}m{#1}}

\begin{document}

\pagestyle{headings}
\mainmatter

\title{SMAP: Single-Shot Multi-Person \\Absolute 3D Pose Estimation} %

\titlerunning{SMAP}

\author{Jianan Zhen\inst{1,2,\star} \and Qi Fang\inst{1,\star} \and Jiaming Sun\inst{1,2} \and Wentao Liu\inst{2} \and \\ Wei Jiang\inst{1} \and Hujun Bao\inst{1} \and Xiaowei Zhou\inst{1}}

\authorrunning{Zhen et al.}

\institute{Zhejiang University \and SenseTime}

\maketitle

\begin{abstract}
Recovering multi-person 3D poses with absolute scales from a single RGB image is a challenging problem due to the inherent depth and scale ambiguity from a single view. Addressing this ambiguity requires to aggregate various cues over the entire image, such as body sizes, scene layouts, and inter-person relationships. However, most previous methods adopt a top-down scheme that first performs 2D pose detection and then regresses the 3D pose and scale for each detected person individually, ignoring global contextual cues.  
In this paper, we propose a novel system that first regresses a set of 2.5D representations of body parts and then reconstructs the 3D absolute poses based on these 2.5D representations with a depth-aware part association algorithm. Such a single-shot bottom-up scheme allows the system to better learn and reason about the inter-person depth relationship, improving both 3D and 2D pose estimation. The experiments demonstrate that the proposed approach achieves the state-of-the-art performance on the CMU Panoptic and MuPoTS-3D datasets and is applicable to in-the-wild videos. {\renewcommand{\thefootnote}{\fnsymbol{footnote}} \footnotetext[1]{Equal contribution. }}

\end{abstract}

\section{Introduction}

Recent years have witnessed an increasing trend of research on monocular 3D human pose estimation because of its wide applications in augmented reality, human-computer interaction, and video analysis. This paper aims to address the problem of estimating absolute 3D poses of multiple people simultaneously from a single RGB image. Compared to the single-person 3D pose estimation problem that focuses on recovering the root-relative pose, i.e., the 3D locations of human-body keypoints relative to the root of the skeleton, the task addressed here additionally needs to recover the 3D translation of each person in the camera coordinate system.

\begin{figure}[t]
	\centering
	\includegraphics[width=0.8\linewidth,trim={7cm 2.2cm 7cm 2cm},clip]{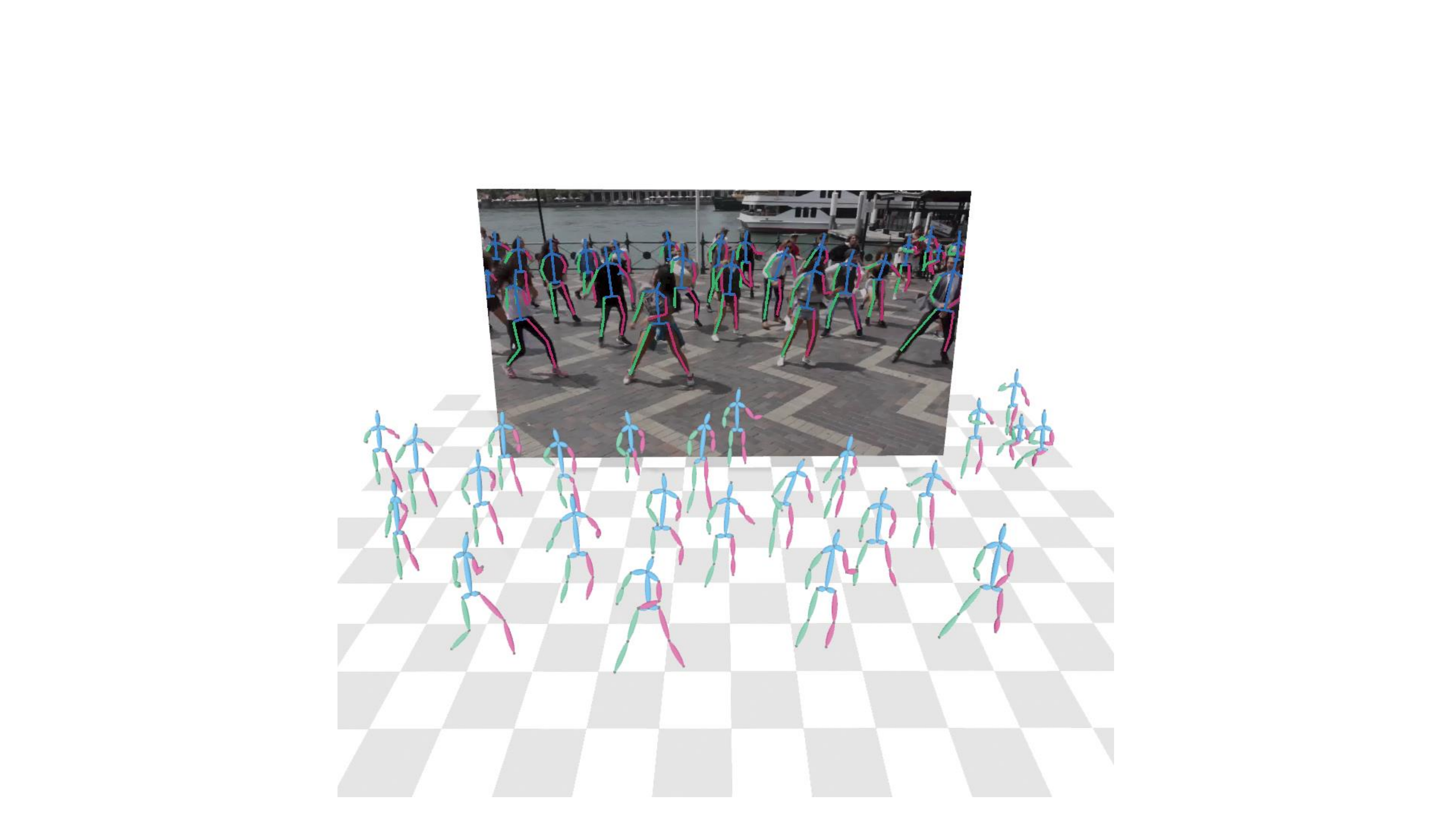}
	\caption{
		We propose a novel framework named SMAP to estimate absolute 3D poses of multiple people from a single RGB image. The figure visualizes the result of SMAP on an in-the-wild image. The proposed single-shot and bottom-up design allows SMAP to leverage the entire image to infer the absolute locations of multiple people consistently, especially in terms of the ordinal depth relations. 
	}
	\label{fig:demo}
\end{figure}

While there has been remarkable progress in recovering the root-relative 3D pose of a single person from an image \cite{sun2018integral,martinez2017simple,Kanazawa:CVPR:2018,guler2019holopose}, it was not until recently that more attention was paid to the multi-person case. 
Most existing methods for multi-person 3D pose estimation extend the single-person approach with a separate stage to recover the absolute position of each detected person separately. 
They either use another neural network to regress the 3D translation of the person from the cropped image \cite{Moon_2019_ICCV_3DMPPE} or compute it based on the prior about the body size \cite{dabral2019multi,zanfir2018monocular,zanfir2018deep}, which ignore the global context of the whole image. Another line of work tries to recover body positions with a ground plane constraint \cite{mehta2019xnect}, but this approach assumes that the feet are visible, which is not always true, and accurate estimation of ground plane geometry from a single image is still an open problem.

We argue that the robust estimation of global positions of human bodies requires to aggregate the depth-related cues over the whole image, such as the 2D sizes of human bodies, the occlusion between them, and the layout of the scene. Recent advances in monocular depth estimation have shown that convolutional neural networks (CNNs) are able to predict the depth map from an RGB image \cite{li2018megadepth,lee2019monocular}, which is particularly successful on human images \cite{Li_2019_CVPR}. This observation motivates us to directly learn the depths of human bodies from the input image instead of recovering them in a post-processing stage. 

To this end, we propose a novel single-shot bottom-up approach to multi-person 3D pose estimation, which predicts absolute 3D positions and poses of multiple people in a single forward pass. We regress the root depths of human bodies in the form of a novel root depth map, which only requires 3D pose annotations as supervision. We train a fully convolutional network to regress the root depth map, as well as 2D keypoint heatmaps, part affinity fields (PAFs), and part relative-depth maps that encode the relative depth between two joints of each body part. Then, the detected 2D keypoints are grouped into individuals based on PAFs using a part association algorithm,  and absolute 3D poses are recovered with the root depth map and part relative-depth maps. The whole pipeline is illustrated in Fig.~\ref{fig:pipeline}. 

We also show that predicting depths of human bodies is beneficial for the part association and 2D pose estimation. We observe that many association errors occur when two human bodies overlap in the image. Knowing the depths of them allows us to reason about the occlusion between them when assigning the detected keypoints. Moreover, from the estimated depth, we can infer the spatial extent of each person in 2D and avoid linking two keypoints with an unreasonable distance. With these considerations, we propose a novel depth-aware part association algorithm and experimentally demonstrate its effectiveness.

To summarize, the contributions of this work are:
\begin{itemize}

\item A single-shot bottom-up framework for multi-person 3D pose estimation, which can reliably estimate absolute positions of multiple people by leveraging depth-relevant cues over the entire image.

\item A depth-aware part association algorithm to reason about inter-person occlusion and bone-length constraints based on predicted body depths, which also benefits 2D pose estimation.

\item The state-of-the-art performance on public benchmarks, with the generalization to in-the-wild images and the flexibility for both whole-body and half-body pose estimation. The code, demonstration videos and other supplementary material are available at \url{https://zju3dv.github.io/SMAP}.
\end{itemize}

\section{Related work}

\paragraph{\bf Multi-person 2D pose.}
Existing methods for multi-person 2D pose estimation can be approximately divided into two classes. Top-down approaches detect human first and then estimate keypoints with a single person pose estimator \cite{chen2018cascaded,fang2017rmpe,papandreou2017towards,xiao2018simple}. Bottom-up approaches localize all keypoints in the image first and then group them to individuals \cite{cao2017realtime,2019arXiv190913423H,newell2017associative,pishchulin2016deepcut,insafutdinov2016deepercut,papandreou2018personlab,nie2019single}. Cao et al.~\cite{cao2017realtime} propose OpenPose and use part affinity fields (PAFs) to represent the connection confidence between keypoints. They solve the part association problem with a greedy strategy. Newell et al.~\cite{newell2017associative} propose an approach that simultaneously outputs detection and group assignments in the form of pixel-wise tags. 

\paragraph{\bf Single-person 3D pose.}
Researches on single-person 3D pose estimation from a single image have already achieved remarkable performances in recent years. One-stage approaches directly regress 3D keypoints from images and can leverage shading and occlusion information to resolve the depth ambiguity. Most of them are learning-based \cite{alp2018densepose,sun2018integral,pavlakos2018ordinal,pavlakos2017coarse,sun2017compositional,yang20183d,guler2019holopose,mehta2017vnect}. Two-stage approaches estimate the 2D pose first and then lift it to the 3D pose, including learning-based\cite{martinez2017simple,pavllo20193d,zhao2019semantic}, optimization-based \cite{zhou2016sparseness,xiang2019monocular} and exemplar-based \cite{chen20173d} methods, which can benefit from the reliable result of 2D pose estimation. 

\paragraph{\bf Multi-person 3D pose.}
For the task of multi-person 3D pose estimation, top-down approaches focus on how to integrate the pose estimation task with the detection framework. They crop the image first and then regress the 3D pose with a single-person 3D pose estimator. Most of them estimate the translation of each person with an optimization strategy that minimizes the reprojection error computed over sparse keypoints \cite{dabral2019multi,rogez2017lcr,rogez2019lcr} or dense semantic correspondences \cite{zanfir2018monocular}. Moon et al. \cite{Moon_2019_ICCV_3DMPPE} regard the area of 2D bounding box as a prior and adopt a neural network to learn a correction factor. In their framework, they regress the root-relative pose and the root depth separately. Informative cues for inferring the interaction between people may lose during the cropping operation. Another work \cite{veges2019absolute} regresses the full depth map based on the existing depth estimation framework \cite{li2018megadepth}, but their `read-out' strategy is not robust to 2D outliers. 
On the other hand, bottom-up approaches focus on how to represent pose annotations as several maps in a robust way \cite{mehta2018single,benzine2019deep,zanfir2018deep,mehta2019xnect}. However, they either optimize the translation in a post-processing way or ignore the 
task of root localization. XNect \cite{mehta2019xnect} extends the 2D location map in~\cite{mehta2017vnect} to 3D ones and estimates the translation with a calibrated camera and the ground plane constraint, but the feet may be invisible in crowded scenes and obtaining the extrinsic parameters of the camera is not practical in most applications. Another line of work tries to recover the SMPL model \cite{zanfir2018deep,zanfir2018monocular}, and their focus lies in using scene constraints and avoiding interpenetration, which is weakly related to our task. Taking these factors into consideration, a framework that both considers recovering the translation in a single forward pass and aggregating global features over the image will be helpful to this task.

\paragraph{\bf Monocular depth estimation.} 
Depth estimation from a single view suffers from inherent ambiguity. Nevertheless, several methods make remarkable advances in recent years \cite{li2018megadepth,lee2019monocular}. Li et al. \cite{Li_2019_CVPR} observe that Mannequin Challenge could be a good source for human depth datasets. They generate training data using multi-view stereo reconstruction and adopt a data-driven approach to recover a dense depth map, achieving good results. However, such a depth map lacks scale consistency and cannot reflect the real depth.

\section{Methods}
\begin{figure*}[t]
	\centering
	\includegraphics[width=0.99\linewidth,trim={0cm 1cm 3.5cm 0cm},clip]{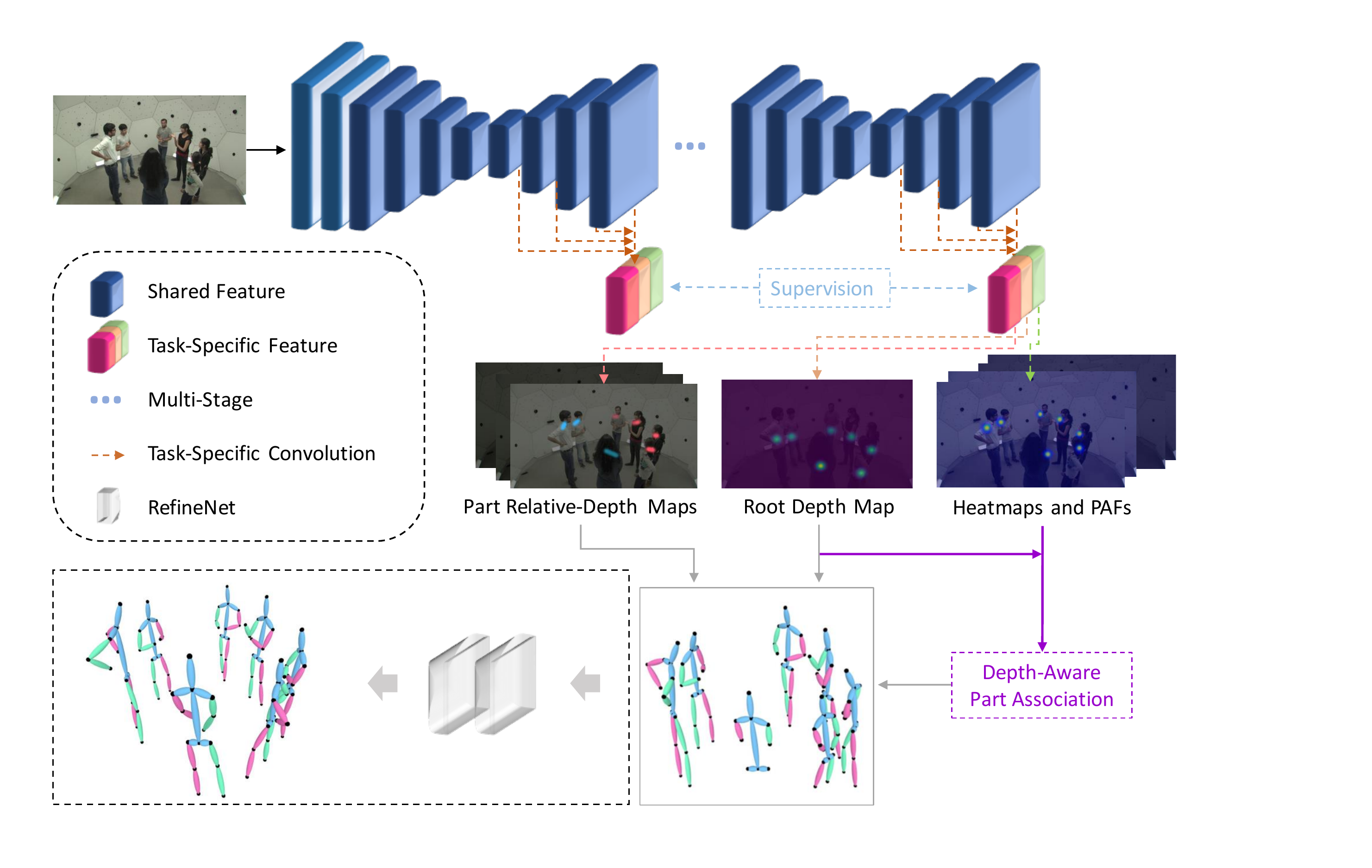}
 	\caption{
 	\textbf{Overview of the proposed approach}. Given a single image, our single-shot network SMAP regresses several intermediate representations including 2D keypoint heatmaps, part affinity fields (PAFs), root depth map, and part relative-depth maps (Red means the child joint has a larger depth than its parent joint, and blue means the opposite). With a new depth-aware part association algorithm, body parts belonging to the same person are linked. With all these intermediate representations combined, absolute 3D poses of all people can be recovered. Finally, an optional RefineNet can be used to further refine the recovered 3D poses and complete invisible keypoints. 
 	}
 	\label{fig:pipeline}
 \end{figure*}

Fig.~\ref{fig:pipeline} presents the pipeline of our approach, which consists of a single-shot bottom-up framework named SMAP. With a single RGB image as input, SMAP outputs 2D representations including keypoint heatmaps and part affinity fields \cite{cao2017realtime}. Additionally, it also regresses 2.5D representations including root depth map and part relative-depth maps, which encode depth information of human bodies. Then, a depth-aware part association algorithm is proposed to assign detected 2D keypoints to individuals, depending on an ordinal prior and an adaptive bone-length constraint. Based on these results, the absolute 3D pose of each person can be reconstructed with a camera model. Individual modules of our system are introduced below.

\subsection{Intermediate representations}\label{sec:representation}
Given the input image, SMAP regresses the following intermediate representations, based on which 3D poses will be reconstructed:

\paragraph{\bf Root depth map.} As the number of people in an input image is unknown, we propose to represent the absolute depths for all human bodies in the image by a novel root depth map. The root depth map has the same size as the input image. The map values at the 2D locations of root joints of skeletons equal their absolute depths. An example is shown in Fig.~\ref{fig:pipeline}. In this way, we are able to represent the depths of multiple people without predefining the number of people. During training, we only supervise the values of root locations. The proposed root depth map can be learned together with other cues within the same network as shown in Fig.~\ref{fig:pipeline} and only requires 3D poses (instead of full depth maps) as supervision, making our algorithm very efficient in terms of both model complexity and training data.

It is worth noting that visual perception of object scale and depth depends on the size of field of view (FoV), i.e., the ratio between the image size and the focal length. 
If two images are obtained with different FoVs, the same person at the same depth will occupy different proportions in these two images and seem to have different depths for the neural network, which may mislead the learning of depth. Thus, we normalize the root depth by the size of FoV as follows:
\begin{equation}
    \widetilde{Z} = Z\frac{w}{f},
\end{equation}
where $\widetilde{Z}$ is the normalized depth, $Z$ is the original depth, and $f$ and $w$ are the focal length and the image width both in pixels, respectively. So $w/f$ is irrelevant to image resolution, but equals to the ratio between the physical size of image sensor and the focal length both in millimeters, i.e., FoV. The normalized depth values can be converted back to metric values in inference. 

\paragraph{\bf Keypoint heatmaps.} Each keypoint heatmap indicates the probable locations of a specific type of keypoints for all people in the image. Gaussian distribution is used to model uncertainties at the corresponding location.

\paragraph{\bf Part affinity fields (PAFs).} PAFs proposed in \cite{cao2017realtime} include a set of 2D vector fields. Each vector field corresponds to a type of body part where the vector at each pixel represents the 2D orientation of the corresponding body part. 

\paragraph{\bf Part relative-depth map.} Besides the root depth, we also need depth values of other keypoints to reconstruct a 3D pose. Instead of predicting absolute depth values, for other keypoints we only regress their relative depths compared to their parent nodes in the skeleton, which are represented by part relative-depth maps. Similar to PAFs, each part relative-depth map corresponds to a type of body part, and every pixel that belongs to a body part encodes the relative depth between two joints of the corresponding body part. This dense representation provides rich information to reconstruct a 3D skeleton even if some keypoints are invisible. 

\paragraph{\bf Network architecture.} 
We use Hourglass \cite{newell2016stacked} as our backbone and modify it to a multi-task structure with multiple branches that simultaneously output the above representations as illustrated in Fig.~\ref{fig:pipeline}. Suppose the number of predefined joints is $J$. Then, there are $4J-2$ channels in total (heatmaps and PAFs: $J+2(J-1)$, root depth map: $1$, part relative-depth map: $J-1$). Each output branch in our network only consists of two convolutional layers. Inspired by \cite{li2019rethinking}, we adopt multi-scale intermediate supervision. The $L_1$ loss is used to supervise the root depths and $L_2$ losses on other outputs. The effect of the network size and the multi-scale supervision will be validated in the experiments.

\subsection{Depth-aware part association}

\begin{figure}[t]
	\centering
	\includegraphics[width=0.65\linewidth,trim={1.6cm 4.7cm 1.6cm 6.0cm},clip]{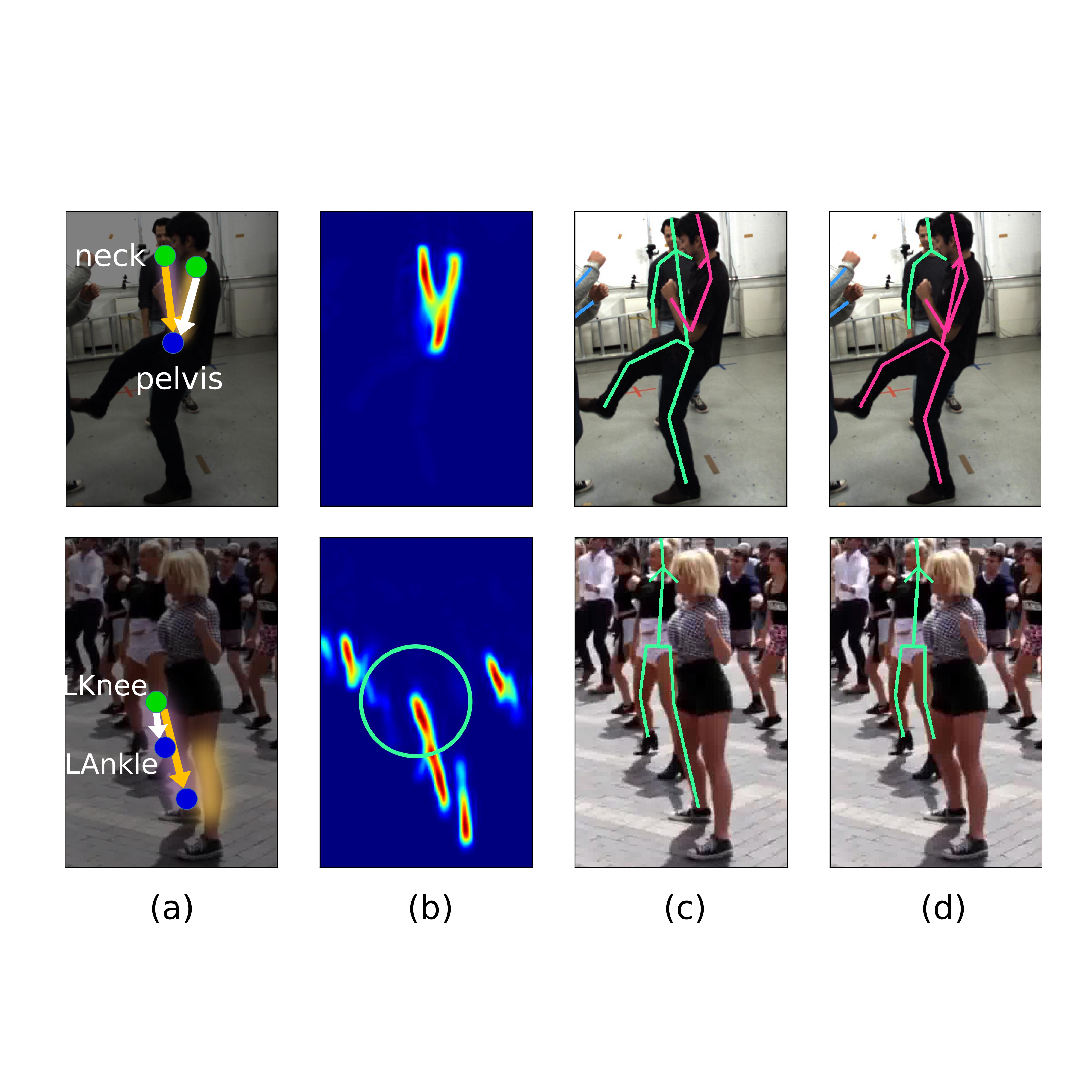}
 	\caption{
 		\textbf{Depth-aware part association}. From left to right: candidate links, part affinity fields, pose estimation results from \cite{cao2017realtime}, and our results with depth-aware part association. 
 		 The example in the first row shows the effect of ordinal prior. Under this circumstance, \cite{cao2017realtime} assigns the pelvis to the occluded person while we give priority to the front person. The example in the second row shows the effect of the adaptive bone-length threshold, indicated by the green circle. As a noisy response occurs at the right ankle of the person it doesn't belong to, \cite{cao2017realtime} will induce a false connection while our algorithm will not.
 	}
 	\label{fig:2DAmbiguity}
 \end{figure}
 
Given 2D coordinates of keypoints from keypoint heatmaps after non-maximum suppression, we need to associate detected joints with corresponding individuals. Cao et al.~\cite{cao2017realtime} propose to link joints greedily based on the association scores given by PAFs. Basically, we follow their method to calculate association scores. However, PAFs scores might be unreliable due to occlusion between people. Fig.~\ref{fig:2DAmbiguity} shows two examples where the above association strategy fails. 

We propose to leverage the estimated depth maps to address the part association ambiguities caused by inter-person occlusion. 

\paragraph{\bf Ordinal prior.} A key insight to solve the occlusion issue is to give priority to the unoccluded person when assigning joints. The occlusion status can be inferred from the depth map. Therefore, we sort root joints from near to far according to the predicted root depth, instead of following the order of PAFs scores. Our association process starts with the root and proceeds along the skeleton tree successively. 

\paragraph{\bf Adaptive bone-length constraint.} To avoid linking two keypoints with an unreasonable distance, Cao et al.~\cite{cao2017realtime} constrain the association with a threshold defined as half of the image size. However, such a fixed threshold is ineffective as the 2D bone length depends on its depth. We adopt an adaptive bone-length threshold to penalize the unreasonable association, depending on predicted body depths. For each part, we compute the mean bone length in the training set in advance. Then, its maximal length in 2D is computed and used as the distance threshold for the corresponding part:

\begin{equation}\label{eq:adaptive}
	d_{cons} = \lambda \cdot \frac{D_{bone} \cdot f}{Z \cdot w} = \lambda \cdot \frac{D_{bone}}{\widetilde{Z}}~,
\end{equation}
where $D_{bone}$ is the 3D average length of a limb, $\widetilde{Z}$ is the normalized root depth predicted by our network, and $\lambda$ is a relaxation factor. $d_{cons}$ is used to filter unreasonable links. From Eq.~\ref{eq:adaptive} we can see that the adaptive threshold is not affected by camera intrinsic parameters or image resizing and is only related to the depth value estimated by our network and the statistical bone length. 

As the association can benefit from depth information, we call it depth-aware part association. Fig.~\ref{fig:2DAmbiguity}(d) shows our qualitative results. The proposed scheme will also be validated by experiments.

\subsection{3D pose reconstruction}

\paragraph{\bf Reconstruction.} Following the connection relations obtained from part association, relative depths from child nodes to parent nodes can be read from the corresponding locations of part relative-depth maps. With the root depth and relative depths of body parts, we are able to compute the depth of each joint. Given 2D coordinates and joint depths, the 3D pose can be recovered through the perspective camera model:
\begin{equation}\label{eq:perspective}
\left[
\begin{matrix}
X, & Y, & Z
\end{matrix}
\right]^T = Z  K^{-1} 
\left[
\begin{matrix}
x, & y, & 1
\end{matrix}
\right]^T~,
\end{equation}
where $\left[X,Y,Z\right]^T$ and $\left[x,y\right]^T$ are 3D and 2D coordinates of a joint respectively. $K$ is the camera intrinsic matrix, 
which is available in most applications, e.g., from device specifications. In our experiments, we use the focal lengths provided by the datasets (same as \cite{Moon_2019_ICCV_3DMPPE}). For internet images with unknown focal lengths, we use a default value which equals the input image width in pixels. Note that the focal length will not affect the predicted ordinal depth relations between people.

\paragraph{\bf Refinement.}
The above reconstruction procedure may introduce two types of errors. One is the cumulative error in the process of joint localization due to the hierarchical skeleton structure, and the other is caused by back projection when the depth is not accurate enough to calculate X and Y coordinates of 3D pose. Moreover, severe occlusion and truncation frequently occur in crowded scenes, which make some keypoints invisible. Therefore, we use an additional neural network named RefineNet to refine visible keypoints and complete invisible keypoints for each 3D pose. RefineNet consists of five fully connected layers. The inputs are 2D pose and 3D root-relative pose while the output is the refined 3D root-relative pose. The coordinates of invisible keypoints in the input are set to be zero.
Note that RefineNet doesn't change the root depths.

\section{Experiments}

We evaluate the proposed approach on two widely-used datasets and compare it to previous approaches. Besides, we provide thorough ablation analysis to validate our designs.

\subsection{Datasets}
\paragraph{\bf CMU Panoptic}~\cite{Joo_2017_TPAMI} is a large-scale dataset that contains various indoor social activities, captured by multiple cameras. Mutual occlusion between individuals and truncation makes it challenging to recover 3D poses. Following \cite{zanfir2018monocular}, we choose two cameras (16 and 30), 9600 images from four activities (Haggling, Mafia, Ultimatum, Pizza) as our test set, and 160k images from different sequences as our training set.

\paragraph{\bf MuCo-3DHP and MuPoTS-3D}~\cite{mehta2018single}. MuCo-3DHP is an indoor multi-person dataset for training, which is composited from single-person datasets. MuPoTS-3D is a test set consisting of indoor and outdoor scenes with various camera poses, making it a convincing benchmark to test the generalization ability. 

\subsection{Implementation details}
We adopt Adam as optimizer with 2e-4 learning rate, and train two models for 20 epochs on the CMU Panoptic and MuCo-3DHP datasets separately, mixed with COCO data~\cite{lin2014microsoft}. The batch size is 32 and 50\% data in each mini-batch is from COCO (same as \cite{mehta2018single,Moon_2019_ICCV_3DMPPE}). Images are resized to a fixed size 832$\times$512 as the input to the network. Note that resizing doesn't change FoV. Since the COCO dataset lacks 3D pose annotations, weights of 3D losses are set to zero when the COCO data is fed. 

\subsection{Evaluation metrics}
\paragraph{\bf MPJPE.} MPJPE measures the accuracy of the 3D root-relative pose. It calculates the Euclidean distance between the predicted and the groundtruth joint locations averaged over all joints.

\paragraph{\bf RtError.} Root Error (RtError) is defined as the Euclidean distance between the predicted and the groundtruth root locations. 

\paragraph{\bf 3DPCK.} 3DPCK is the percentage of correct keypoints. A keypoint is declared correct if the Euclidean distance between predicted and groundtruth coordinates is smaller than a threshold (15cm in our experiments). PCK$_{\rm rel}$ measures relative pose accuracy with root alignment; PCK$_{\rm abs}$ measures absolute pose accuracy without root alignment; and PCK$_{\rm root}$ only measures the accuracy of root joints. AUC means the area under curve of 3DPCK over various thresholds. 

\paragraph{\bf PCOD.} We propose a new metric named the percentage of correct ordinal depth (PCOD) relations between people. The insight is that predicting absolute depth from a single view is inherently ill-posed, while consistent ordinal relations between people are more meaningful and suffice many applications. For a pair of people $(i, j)$, we compare their root depths and divide the ordinal depth relation into three classes: closer, farther, and roughly the same (within 30cm). PCOD equals the classification accuracy of predicted ordinal depth relations.

\begin{table}[t]
	\centering
	\setlength\tabcolsep{1.0pt}
	\def\arraystretch{1.0}
	\caption{Results on the Panoptic dataset. For \cite{Moon_2019_ICCV_3DMPPE}, we used the code provided by the authors and trained it on the Panoptic dataset. *The average of \cite{zanfir2018deep} is recalculated following the standard practice in \cite{zanfir2018monocular}, i.e., average over activities.}
	\label{table:panoptic3d}
	\begin{tabular}{L{1.4cm}|L{2.85cm}*{5}{C{1.46cm}}}
	\specialrule{.1em}{.05em}{.05em}
    & \multicolumn{1}{c}{Method} & Haggling & Mafia & Ultim. & Pizza & Average  \\\hline
	\multirow{6}{*}{MPJPE} & ~PoPa et al.~\cite{popa2017deep} & 217.9 & 187.3 & 193.6 & 221.3 & 203.4 \\ 
	& ~Zanfir et al.~\cite{zanfir2018monocular} & 140.0 & 165.9 & 150.7 & 156.0 & 153.4 \\
	& ~Moon et al.~\cite{Moon_2019_ICCV_3DMPPE} & 89.6 & 91.3 & 79.6 & 90.1 & 87.6 \\
    & ~Zanfir et al.~\cite{zanfir2018deep} & 72.4 & 78.8 & 66.8 & 94.3 & 78.1*\\
    & ~\textbf{Ours w/o Refine} & 71.8 & 72.5 & 65.9 & 82.1 & 73.1\\
	& ~\textbf{Ours} & \textbf{63.1} & \textbf{60.3} & \textbf{56.6} & \textbf{67.1} & \textbf{61.8} \\
	\hline
	\multirow{3}{*}{RtError} & ~Zanfir et al.~\cite{zanfir2018monocular} & 257.8 & 409.5 & 301.1 & 294.0 & 315.5 \\ 
	& ~Moon et al.~\cite{Moon_2019_ICCV_3DMPPE} & 160.2 & 151.9 & 177.5 & 127.7 & 154.3 \\
	& ~\textbf{Ours} & \textbf{84.7} & \textbf{87.7} & \textbf{91.2} & \textbf{78.5} & \textbf{85.5} \\
	\hline
    \multirow{2}{*}{PCOD} & ~Moon et al.~\cite{Moon_2019_ICCV_3DMPPE} & 92.3 & 93.7 & 95.2 & 94.2 & 93.9 \\
	& ~\textbf{Ours} & \textbf{97.8} & \textbf{98.5} & \textbf{97.6} & \textbf{99.6} & \textbf{98.4} \\ 
	\specialrule{.1em}{.05em}{.05em}
	\end{tabular}
\end{table}

\begin{table}[t]
	\centering
	\setlength\tabcolsep{1.0pt}
	\caption{Results on the MuPoTS-3D dataset. All numbers are average values over 20 activities.}
	\label{table:mupots_all}
	\begin{tabular}{C{1.15cm}|L{1.8cm}|*{5}{C{1.2cm}}|C{1.16cm}C{1.16cm}}
	\specialrule{.1em}{.05em}{.05em}
    & & \multicolumn{5}{c|}{Matched people} & \multicolumn{2}{c}{All people} \\\hline
    & \multicolumn{1}{l|}{~Method} & PCK$_{\rm rel}$ & PCK$_{\rm abs}$ & PCK$_{\rm root}$ & AUC$_{\rm rel}$ & PCOD & PCK$_{\rm rel}$ & PCK$_{\rm abs}$\\ \hline
    \multirow{4}{*}{\shortstack{top\\down}}
    & ~Rogez.~\cite{rogez2017lcr} & 62.4 & - & - & - & - & 53.8 & -\\
    & ~Rogez.~\cite{rogez2019lcr} & 74.0 & - & - & - & - & 70.6 & -\\
    & ~Dabral.~\cite{dabral2019multi} & 74.2 & - & - & - & - & 71.3 & -\\
    & ~Moon.~\cite{Moon_2019_ICCV_3DMPPE} & \textbf{82.5} & 31.8 & 31.0 & 40.9 & 92.6 & \textbf{81.8} & 31.5 \\
    \hline
    \multirow{3}{*}{\shortstack{bottom\\up}}
    & ~Mehta.~\cite{mehta2018single} & 69.8 & - & - & - & - & 65.0 & - \\
	& ~Mehta.~\cite{mehta2019xnect} & 75.8 & - & - & - & - & 70.4 & - \\
	& ~\textbf{Ours} & \textbf{80.5} & \textbf{38.7} & \textbf{45.5} & \textbf{42.7} & \textbf{97.0} & \textbf{73.5} & \textbf{35.4}\\
	\specialrule{.1em}{.05em}{.05em}
	\end{tabular}
\end{table}

\subsection{Comparison with state-of-the-art methods}
\paragraph{\bf CMU Panoptic.} 
Table~\ref{table:panoptic3d} demonstrates quantitative comparison between state-of-the-art methods and our model. It indicates that our model outperforms previous methods in all metrics by a large margin. In particular, the error on the Pizza sequence decreases significantly compared with the previous work. As the Pizza sequence shares no similarity with the training set, this improvement shows our generalization ability. 

\begin{figure}[t!]
	\centering
	\includegraphics[width=0.8\linewidth,trim={0.2cm 5.5cm 0.2cm 6cm},clip]{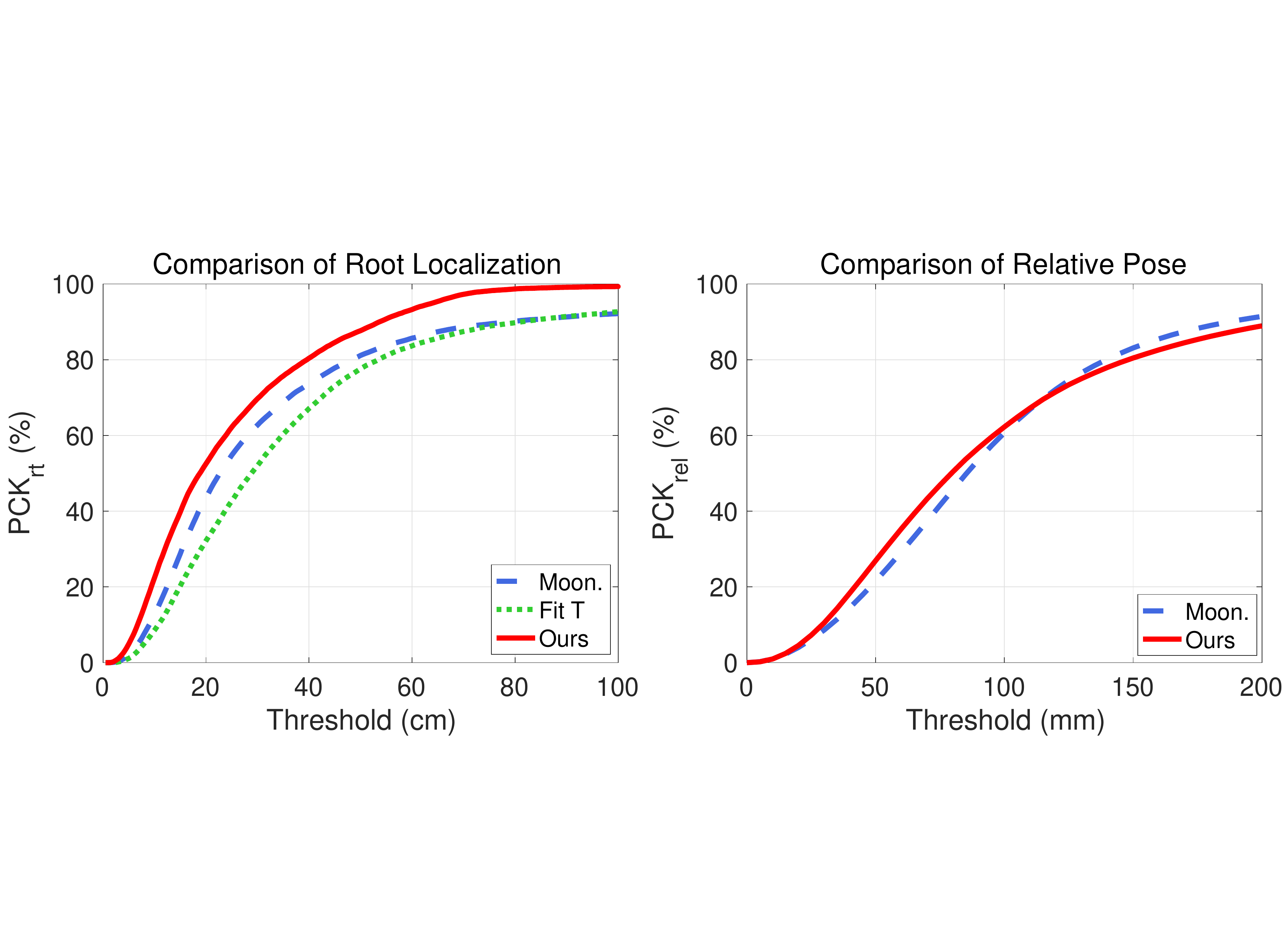}
 	\caption{
 		\textbf{Comparisons of root localization and relative pose}. The curves are PCK$_{\rm root}$ and PCK$_{\rm rel}$ over different thresholds on the MuPoTS-3D dataset. Blue: result of \cite{Moon_2019_ICCV_3DMPPE}. Green: estimating the translation by minimizing the reprojection error. Red: our result.
 	}
 	\label{fig:rootrelPCK}
 \end{figure}

\paragraph{\bf MuPoTS-3D.} 
We follow the protocol of \cite{Moon_2019_ICCV_3DMPPE}. Additionally, PCK$_{\rm root}$ and PCOD are used to evaluate the 3D localization of people. In terms of the absolute pose which we are more concerned with, it can be observed from Table~\ref{table:mupots_all} that our model is superior to \cite{Moon_2019_ICCV_3DMPPE} in relevant metrics including PCK$_{\rm abs}$, PCK$_{\rm root}$ and PCOD by a large margin. It also demonstrates that our model has higher PCK$_{\rm rel}$ compared with all bottom-up methods and most top-down methods except \cite{Moon_2019_ICCV_3DMPPE}. Note that we achieve higher AUC$_{\rm rel}$ compared to \cite{Moon_2019_ICCV_3DMPPE} for the relative 3D pose of matched people.

\begin{figure*}[t!]
	\centering
	\includegraphics[width=0.95\linewidth,trim={1.4cm 4cm 1.3cm 7cm},clip]{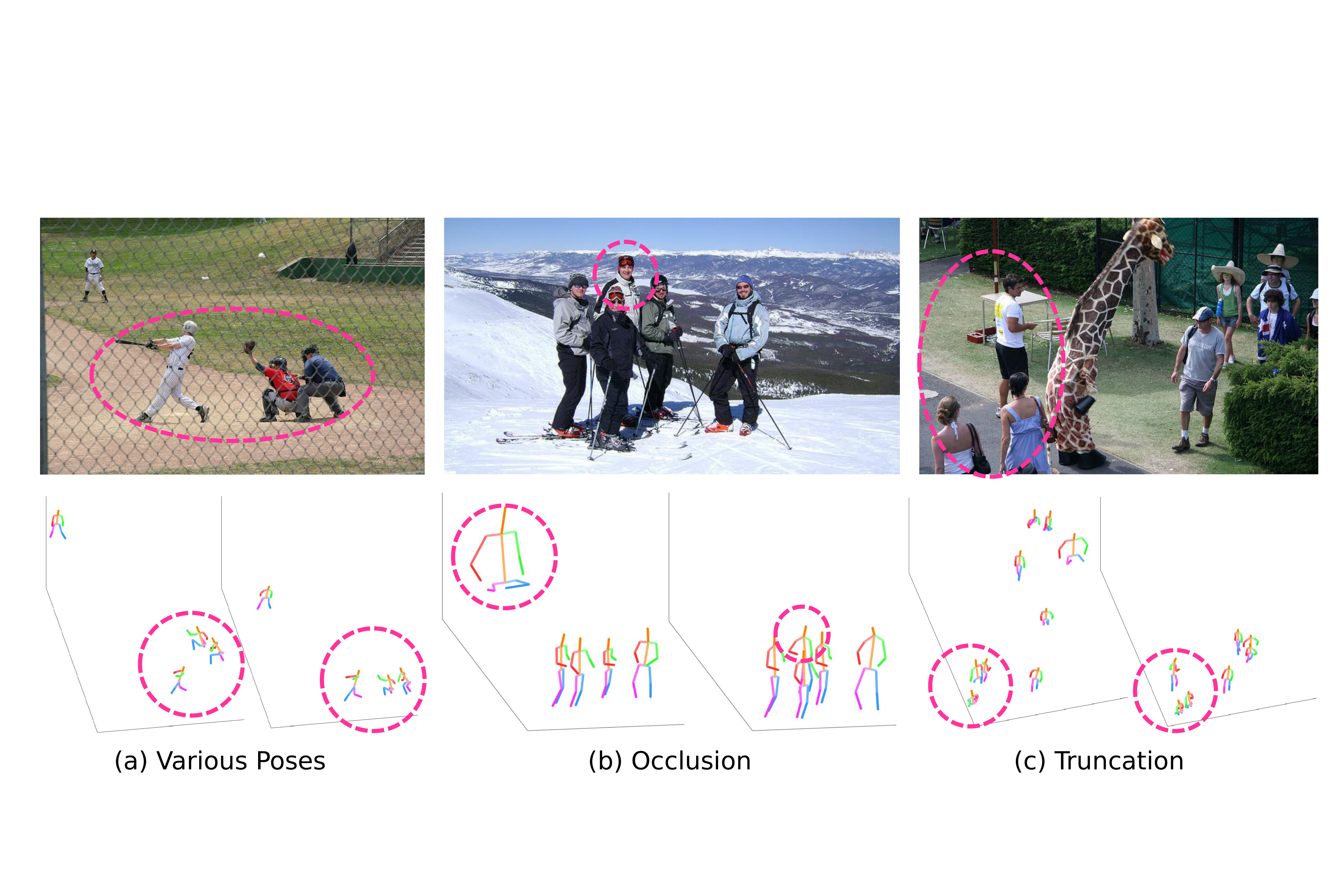}
 	\caption{
		\textbf{Qualitative comparison}. The results of three example images. For each example, the top row shows the input image, and the bottom row shows the results of \cite{Moon_2019_ICCV_3DMPPE} (left) and the proposed method (right), respectively. The red circles highlight the difference in localization of human bodies between two methods.
 	}
 	\label{fig:case}
 \end{figure*}

\paragraph{\bf Comparison with top-down methods.} 
We provide additional analysis to compare our single-shot bottom-up method to the state-of-the-art top-down method \cite{Moon_2019_ICCV_3DMPPE}. Fig.~\ref{fig:rootrelPCK} shows thorough comparisons in terms of PCK$_{\rm root}$ and PCK$_{\rm rel}$. For root localization, we compare to two methods: 1) regressing the scale from each cropped bounding box using a neural network as in \cite{Moon_2019_ICCV_3DMPPE} and 2) estimating the 3D translation by optimizing reprojection error with the groundtruth 2D pose and the estimated relative 3D pose from \cite{Moon_2019_ICCV_3DMPPE} (`FitT' in Fig.~\ref{fig:rootrelPCK}). We achieve better PCK$_{\rm root}$ over various thresholds than both of them. Notably, we achieve roughly 100\% accuracy with a threshold 1m. As for relative pose estimation, \cite{Moon_2019_ICCV_3DMPPE} achieves higher PCK$_{\rm rel}$ (@15cm) as it adopts a separate off-the-shelf network \cite{sun2018integral} that is particularly optimized for relative 3D pose estimation. Despite that, we obtain better PCK$_{\rm rel}$ when the threshold is smaller and higher AUC$_{\rm rel}$. 

Fig.~\ref{fig:case} shows several scenarios (various poses, occlusion, and truncation) in which the top-down method \cite{Moon_2019_ICCV_3DMPPE} may fail as it predicts the scale for each detected person separately and ignore global context. Instead, the proposed bottom-up design is able to leverage features over the entire image instead of only using cropped features in individual bounding boxes.

Furthermore, our running time and memory remain almost unchanged with the number of people in the image while those of \cite{Moon_2019_ICCV_3DMPPE} grow faster with the number of people due to its top-down design, as shown in the supplementary material.

\begin{figure}[t]
\centering
\includegraphics[width=0.9\linewidth,trim={0cm 0cm 0cm 0cm},clip]{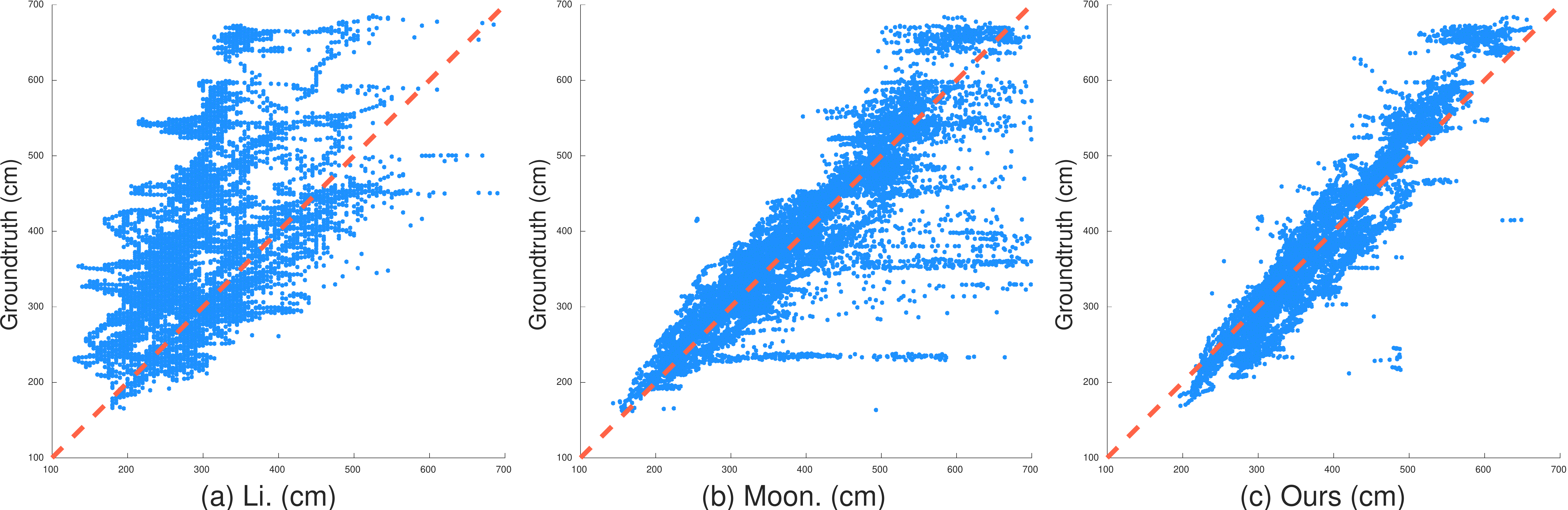}
 \caption{
	 \textbf{Comparison with alternative depth estimation methods}. The scatter plots show the consistency of root depth estimation on the MuPoTS-3D dataset. The X and Y axes are the predicted, groundtruth root depth, respectively. The dashed line means the ideal result, i.e., estimation equals ground truth. (a) `Read-out' root depths from the full depth map estimated by \cite{Li_2019_CVPR}. 
	 (b) State-of-the-art top-down approach \cite{Moon_2019_ICCV_3DMPPE}. 
	 (c) Our approach.
 }
 \label{fig:consistency}
\end{figure}

\paragraph{\bf Depth estimation.} Apart from our method, there are two alternatives for depth estimation: 1) regressing the full depth map rather than the root depth map. 2) using the cropped image as the input to the network rather than the whole image. For the first alternative, since there is no depth map annotation in existing multi-person outdoor datasets, we use the released model of the state-of-the-art human depth estimator \cite{Li_2019_CVPR}, which is particularly optimized for human depth estimation trained on a massive amount of in-the-wild `frozen people' videos. For the second alternative, \cite{Moon_2019_ICCV_3DMPPE} is the state-of-the-art method that estimates root depth from the cropped image, so we compare with it.
Fig.~\ref{fig:consistency} demonstrates scatter plots of the groundtruth root depth versus the predicted root depth of three methods on the MuPoTS-3D dataset. Ideally, the estimated depths should be linearly correlated to the ground truth, resulting a straight line in the scatter plot. Our model shows better consistency than baselines. Note that, while the compared methods are trained on different datasets, the images in the test set MuPoTS-3D are very different from the training images for all methods. Though not rigorous, this comparison is still reasonable to indicate the performance of these methods when applied to unseen images. 

\subsection{Ablation analysis}

\begin{table}[t]
	\centering
	\setlength\tabcolsep{1.0pt}
	\def\arraystretch{1.0}
	\caption{Ablation study of the structure design on the MuPoTS-3D dataset. The default backbone is Hourglass model with three stages, and `Smaller Backbone' means one-stage model. }
	\label{table:ablation}
	\begin{tabular}{L{4.6cm}|*{5}{C{1.2cm}}}
	\specialrule{.1em}{.05em}{.05em}
    Design & Recall & PCK$_{\rm root}$ & PCK$_{\rm abs}$ & PCK$_{\rm rel}$ & PCOD\\\hline
    Full Model & \textbf{92.3} & \textbf{45.5} & \textbf{38.7} & \textbf{80.5} & \textbf{97.0} \\
	No Normalization & 92.3 & 5.7 & 8.7 & 78.9 & 95.7 \\
	No Multi-scale Supervision & 92.1 & 45.2 & 36.2 & 75.4 & 93.1 \\
	No RefineNet & 92.3 & 45.5 & 34.7 & 70.9 & 97.0 \\
	Smaller Backbone & 91.1 & 43.8 & 35.1 & 75.7 & 96.4 \\
	\specialrule{.1em}{.05em}{.05em}
	\end{tabular}
\end{table}
\noindent{\bf Architecture.} Table~\ref{table:ablation} shows how different designs of our framework affect the multi-person 3D pose estimation accuracy:
1) the performance of our model will degrade severely without depth normalization. As we discussed in Section \ref{sec:representation}, normalizing depth values by the size of FoV makes depth learning easier. 
2) Multi-scale supervision is beneficial.
3) To show that our performance gain in terms of the absolute 3D pose is mostly attributed to our single-shot bottom-up design rather than the network size, we test with a smaller backbone. The results show that, even with a one-stage hourglass network, our method still achieves higher PCK$_{\rm root}$ and PCK$_{\rm abs}$ than the top-down method  \cite{Moon_2019_ICCV_3DMPPE}.

\begin{table}[t]
	\centering
	\setlength\tabcolsep{1.0pt}
	\def\arraystretch{1.0}
	\caption{\textbf{Ablation study of the part association}. `2DPA' means the 2D part association proposed by \cite{cao2017realtime}. `DAPA' means the depth-aware part association we proposed. Both of them are based on the same heatmaps and PAFs results.}
	\label{table:partassociation}
	\begin{tabular}{L{1.0cm}|C{1.1cm}C{1.15cm}|*{5}{C{1.2cm}}}
	\specialrule{.1em}{.05em}{.05em}
     & \multicolumn{2}{c|}{Panoptic} & \multicolumn{5}{c}{MuPoTS-3D}   \\\hline
     & Recall & 2DPCK & Recall & PCK$_{\rm root}$ & PCK$_{\rm abs}$ & PCK$_{\rm rel}$ & PCOD \\
     \hline
     2DPA & 94.3 & 92.4 & 92.1 & 45.3 & 38.6 & 80.2 & 96.5\\
     DAPA & \textbf{96.4} & \textbf{93.1} & \textbf{92.3} & \textbf{45.5} & \textbf{38.7} & \textbf{80.5} & \textbf{97.0}\\
	\specialrule{.1em}{.05em}{.05em}
	\end{tabular}
\end{table}

\begin{figure*}[t]
\centering
\includegraphics[width=1.0\linewidth,trim={1.5cm 1cm 1.5cm 0cm},clip]{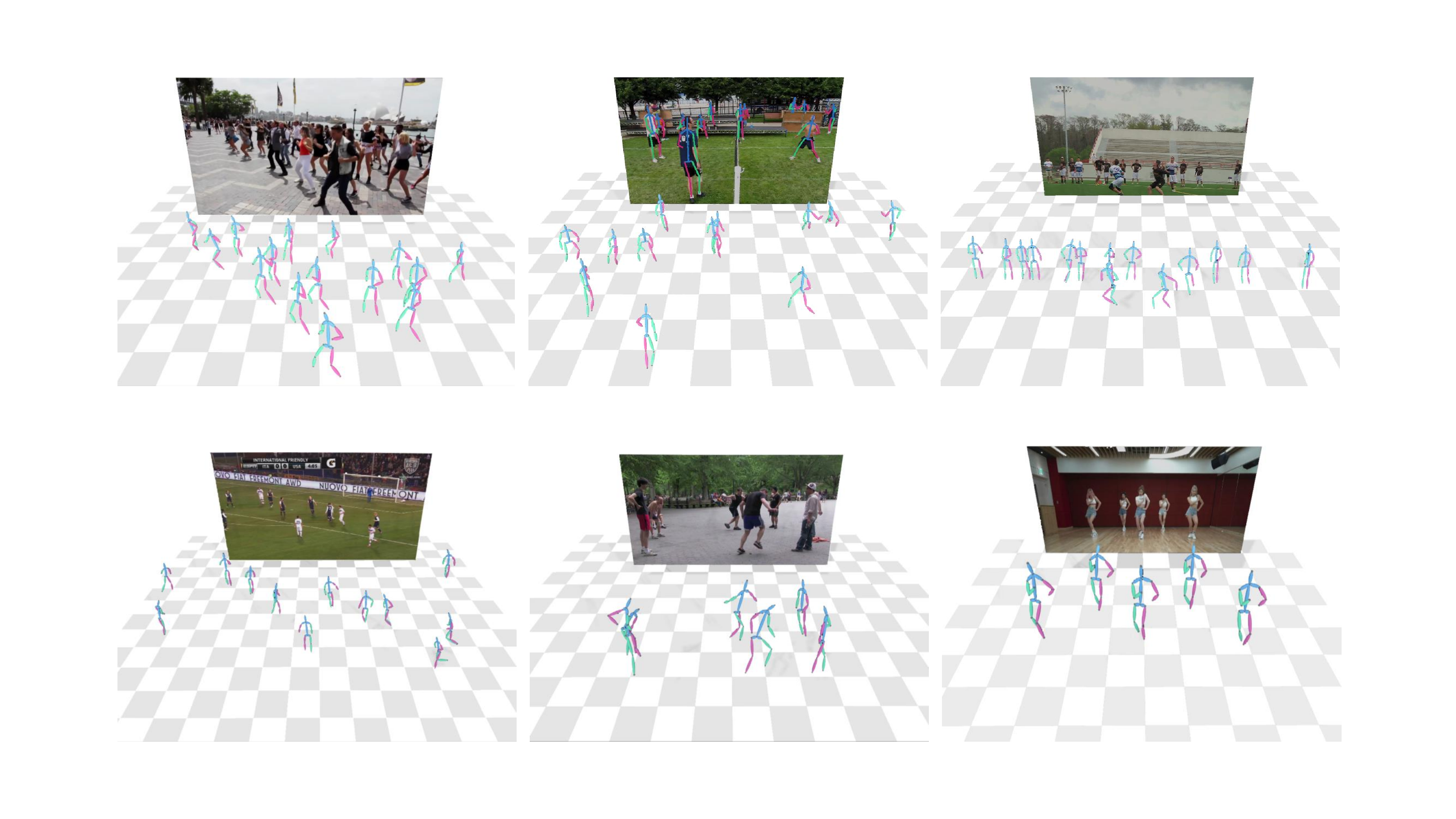} \caption{
	\textbf{Qualitative results on in-the-wild images from the Internet}.
 }
 \label{fig:lastdemo}
\end{figure*}

\paragraph{\bf Part association.} To compare the proposed depth-aware part association with the 2D part association in \cite{cao2017realtime}, 
we evaluate relevant metrics on Panoptic and MuPoTS-3D datasets. Note that the threshold of 2DPCK is the half of the head size. Table~\ref{table:partassociation} lists the results (2DPA vs. DAPA) and reveals that our depth-aware part association outperforms the 2D part association in all these metrics. Besides, Fig.~\ref{fig:2DAmbiguity} shows some qualitative examples.

\paragraph{\bf RefineNet.} 
Table~\ref{table:panoptic3d} and ~\ref{table:ablation} show that
RefineNet is able to improve both relative and absolute pose estimation. It is able to complete invisible keypoints and refine visible keypoints with a learned 3D pose prior. 
The improvement is more significant on the Panoptic dataset since the training and test images are captured by cameras with similar views. 

~\\
Please refer to the supplementary material for more experimental details and results.

\section{Conclusion}

We proposed a novel single-shot bottom-up framework to estimate absolute multi-person 3D poses from a single RGB image. The proposed framework uses a fully convolutional network to regress a set of 2.5D representations for multiple people, from which the absolute 3D poses can be reconstructed. Additionally, benefited from the depth estimation of human bodies, a novel depth-aware part association algorithm was proposed and proven to benefit 2D pose estimation in crowd scenes. Experiments demonstrated state-of-the-art performance as well as generalization ability of the proposed approach.

\paragraph{\bf Acknowledgements:} The authors would like to acknowledge support from NSFC (No. 61806176), Fundamental Research Funds for the Central Universities (2019QNA5022) and ZJU-SenseTime Joint Lab of 3D Vision.

\clearpage
\begin{center}
    \textbf{\Large Supplementary Material: \\SMAP: Single-Shot Multi-Person \\Absolute 3D Pose Estimation} 
\end{center}
\setcounter{section}{0}

In this supplementary material, we provide more experimental details and results. Additionally, qualitative results on in-the-wild images from the Internet are shown in the supplementary video.

\section{More details}
\subsection{Loss function}
There are three output branches of the network, illustrated in  Fig.~\ref{fig:pipeline}. The first branch regresses keypoint heatmaps $\bm H_J$ and PAFs $\bm C$ simultaneously, while the second branch regresses part relative-depth maps $\bm H_{\Delta Z}$. $L_2$ loss is applied to these two branches. The third branch predicts root depth map $\bm H_{RZ}$. According to 2D location of detected root $(x^{root}, y^{root})$, we can get the predicted root depth $\bm{H}_{RZ}(x^{root}, y^{root})$ and compared it with the groundtruth normalized depth $\widetilde{Z}^*$ using $L_1$ loss. The total loss is computed by weighted summation of all losses.
Our loss functions are as follows. 
\begin{equation*}\label{eq:loss}
\begin{split}
L_{total} =&~~w_{2D} \cdot L_{2D} + w_{\Delta Z}\cdot L_{\Delta Z} + w_{RZ} \cdot L_{RZ} \\
L_{2D} =& ~\sum_{i=1}^N \sum_p ||\bm{H}_{J,i}(p) - \bm{H}_{J,i}^*(p)||_2^2 ~~+~~\\ 
& \sum_{i=1}^{2N-2} \sum_p ||\bm{C}_i(p) - \bm{C}_i^*(p)||_2^2 \\
L_{\Delta Z} =& ~\sum_{i=1}^{N-1} \sum_p ||\bm{H}_{\Delta Z, i}(p) - \bm{H}_{\Delta Z, i}^*(p)||_2^2 \\
L_{RZ} =& ~~\sum_{i=1}^{M} ||\hm{H}_{RZ}(x^{root}_{i}, y^{root}_{i}) - \widetilde{Z}_i^{*}||_1~,
\end{split}
\end{equation*}
where $N$, $M$ are the number of joints, the number of detected people (root joints) respectively, $p$ means each pixel location and superscript $*$ denotes the groundtruth. The default settings are: $w_{2D}$=0.1, $w_{\Delta Z}$=5, $w_{RZ}$=10. 

\subsection{Running time and memory}

\begin{table}[t]
	\centering
	\setlength\tabcolsep{1.0pt}
	\def\arraystretch{1.0}
	\caption{Running time and memory comparison.}
	\label{table:time_memory}
    \begin{tabular}{C{1cm}C{1.5cm}||C{1.8cm}C{1.8cm}||C{1.8cm}C{1.8cm}}
	\specialrule{.1em}{.05em}{.05em}
	& & \multicolumn{2}{c||}{3-people} & \multicolumn{2}{c}{20-people} \\
	\hline
    & & Time(ms) & Memory(M) & Time(ms) & Memory(M) \\
    \hline
    \multirow{3}{*}{\cite{Moon_2019_ICCV_3DMPPE}} & DetectNet & 120.0 & 899 & 120.0 & 899 \\ 
    & PoseNet & 14.7 & 815 & 71.8 & 1491 \\
    & RootNet & 13.0 & 803 & 58.9 & 1051 \\
    \hline
    \multirow{3}{*}{Ours} & SMAP & 57.0 & 1379 & 57.0 & 1379 \\
    & DAPA & 4.5 & - & 8.8 & - \\
    & RefineNet & 0.80 & $\sim$0.5 & 0.83 & $\sim$0.5 \\
    \specialrule{.1em}{.05em}{.05em}
    \end{tabular}
\end{table}

Table \ref{table:time_memory} provides detailed information about running time and memory of the state-of-the-art top-down method \cite{Moon_2019_ICCV_3DMPPE} and our method. Note that our method is almost not affected by the number of people in the image. 

\section{More results compared with SOTA}
Due to the limited space, only the average PCK$_{\rm abs}$ is reported in the main manuscript. Here we provide more thorough experimental results. Table~\ref{table:mupots_abs20} presents sequence-wise PCK$_{\rm abs}$ on the MuPoTS-3D dataset and demonstrates that our PCK$_{\rm abs}$ is higher than the state-of-the-art top-down method \cite{Moon_2019_ICCV_3DMPPE}, especially for outdoor scenarios (TS6-TS20). Table~\ref{table:mupots_rel} shows that our model has higher PCK$_{\rm rel}$ compared with all bottom-up methods and most top-down methods except \cite{Moon_2019_ICCV_3DMPPE}. Note that we have higher AUC$_{\rm rel}$ compared with \cite{Moon_2019_ICCV_3DMPPE} as we state in the main manuscript. Table \ref{table:h36m} shows the results on the Human3.6M dataset.

\section{More ablation analysis}
\subsection{Effect of the multi-task structure}
SMAP simultaneously output 2D information (keypoint heatmaps and PAFs), root depth map, and part relative-depth map. To analyze the impact of our single-shot multi-task structure on root localization, we delete some of the output branches and evaluate the performance, as indicated in Table~\ref{table:ablation_root}. One variant is only to regress the root position and its depth alone (row 2 of Table~\ref{table:ablation_root}). This variant can obtain an acceptable result, which reflects the significance of our bottom-up design for root localization. Another variant which adds the keypoint heatmaps and PAFs branches (row 3 of Table~\ref{table:ablation_root}) significantly improves the performance, indicating that 2D cues (pose, body size) are also beneficial to root depth estimation. Nevertheless, this variant is still inferior to the full model. 

\subsection{Influence of camera intrinsics}
Here we make three comparisons:1) full model with known camera intrinsics. 2) full model without camera intrinsics. 3) without normalization. 

RtError of our full model reaches 23.3cm on the MuPoTS-3D dataset. If the intrinsic parameter is not provided (use default intrinsics), RtError increases to 67cm. Note that the ordinal depth relation remains unchanged. If the model lacks normalization, RtError is as high as 120cm.

\begin{table}[t]
	\centering
	\setlength\tabcolsep{1.0pt}
	\def\arraystretch{1.0}
	\caption{Sequence-wise PCK$_{\rm abs}$ on the MuPoTS-3D dataset for matched groundtruths.}
	\label{table:mupots_abs20}
	\begin{tabular}{L{2.5cm}*{11}{C{0.73cm}}}
	\specialrule{.1em}{.05em}{.05em}
     & S1  & S2 & S3 & S4 & S5 & S6 & S7 & S8 & S9 & S10 \\ \hline
	Moon et al.~\cite{Moon_2019_ICCV_3DMPPE} & \textbf{59.5} & \textbf{45.3} & \textbf{51.4} & \textbf{46.2} & 53.0 & \textbf{27.4} & 23.7 & \textbf{26.4} & \textbf{39.1} & 23.6 & \\
    \textbf{Ours} & 42.1 & 41.4 & 46.5 & 16.3 & \textbf{53.0} & 26.4 & \textbf{47.5} & 18.7 & 36.7 & \textbf{73.5} &\\
    \specialrule{.1em}{.05em}{.05em}
    & S11 & S12 & S13 & S14 & S15 & S16 & S17 & S18 & S19 & S20 & Avg. \\ \hline
    Moon et al.~\cite{Moon_2019_ICCV_3DMPPE} & 18.3 & 14.9 & \textbf{38.2} & 29.5 & 36.8 & 23.6 & 14.4 & 20.0 & 18.8 & 25.4 & 31.8 \\
	\textbf{Ours} & \textbf{46.0} & \textbf{22.7} & 24.3 & \textbf{38.9} & \textbf{47.5} & \textbf{34.2} & \textbf{35.0} & \textbf{20.0} & \textbf{38.7} & \textbf{64.8} & \textbf{38.7}\\
    
	\specialrule{.1em}{.05em}{.05em}
	\end{tabular}
	\vspace*{-2mm}
\end{table}

\begin{table}[t]
	\centering
	\setlength\tabcolsep{1.0pt}
	\def\arraystretch{1.0}
	\caption{PCK$_{\rm rel}$ on the MuPoTS-3D dataset for matched groundtruths. 
	`T' denotes top-down methods while `B' denotes bottom-up methods.}
	\label{table:mupots_rel}
	\begin{tabular}{L{2.5cm}L{0.5cm} *{11}{C{0.73cm}}}
	\specialrule{.1em}{.05em}{.05em}
    & & S1 & S2 & S3 & S4 & S5 & S6 & S7 & S8 & S9 & S10 \\ \hline
    Rogez et al.~\cite{rogez2017lcr} & T & 69.1 & 67.3 & 54.6 & 61.7 & 74.5 & 25.2 & 48.4 & 63.3 & 69.0 & 78.1 & \\
    Rogez et al.~\cite{rogez2019lcr} & T & 88.0 & 73.3 & 67.9 & 74.6 & 81.8 & 50.1 & 60.6 & 60.8 & 78.2 & 89.5 &  \\
    Dabral et al.~\cite{dabral2019multi} & T & 85.8 & 73.6 & 61.1 & 55.7 & 77.9 & 53.3 & 75.1 & 65.5 & 54.2 & 81.3 & \\
	Moon et al.~\cite{Moon_2019_ICCV_3DMPPE} & T & \textbf{94.4} & 78.6 & \textbf{79.0} & \textbf{82.1} & 86.6 & \textbf{72.8} & 81.9 & 75.8 & \textbf{90.2} & \textbf{90.4} & \\
	Mehta et al.~\cite{mehta2018single} & B & 81.0 & 64.3 & 64.6 & 63.7 & 73.8 & 30.3 & 65.1 & 60.7 & 64.1 & 83.9 &  \\
	Mehta et al.~\cite{mehta2019xnect} & B & 88.4 & 70.4 & 68.3 & 73.6 & 82.4 & 46.4 & 66.1 & \textbf{83.4} & 75.1 & 82.4 &  \\
    \textbf{Ours} & B & 89.9 & \textbf{88.3} & 78.9 & 78.2 & \textbf{87.6} & 51.0 & \textbf{88.5} & 71.6 & 70.3 & 89.2 &\\
    \specialrule{.1em}{.05em}{.05em}
    & & S11 & S12 & S13 & S14 & S15 & S16 & S17 & S18 & S19 & S20 & Avg. \\ \hline
    Rogez et al.~\cite{rogez2017lcr} & T & 53.8 & 52.2 & 60.5 & 60.9 & 59.1 & 70.5 & 76.0 & 70.0 & 77.1 & 81.4 & 62.4 \\
    Rogez et al.~\cite{rogez2019lcr} & T & 70.8 & 74.4 & 72.8 & 64.5 & 74.2 & 84.9 & 85.2 & 78.4 & 75.8 & 74.4 & 74.0 \\
    Dabral et al.~\cite{dabral2019multi} & T & \textbf{82.2} & 71.0 & 70.1 & 67.7 & 69.9 & 90.5 & 85.7 & \textbf{86.3} & 85.0 & \textbf{91.4} & 74.2 \\
    Moon et al.~\cite{Moon_2019_ICCV_3DMPPE} & T & 79.4 & 79.9 & \textbf{75.3} & \textbf{81.0} & \textbf{81.0} & 90.7 & 89.6 & 83.1 & 81.7 & 77.3 & \textbf{82.5} \\
    Mehta et al.~\cite{mehta2018single} & B & 71.5 & 69.6 & 69.0 & 69.6 & 71.1 & 82.9 & 79.6 & 72.2 & 76.2 & 85.9 & 69.8 \\
	Mehta et al.~\cite{mehta2019xnect} & B & 76.5 & 73.0 & 72.4 & 73.8 & 74.0 & 83.6 & 84.3 & 73.9 & \textbf{85.7} & 90.6 & 75.8 \\
	\textbf{Ours} & B & 76.3 & \textbf{82.0} & 70.8 & 65.2 & 80.4 & \textbf{91.6} & \textbf{90.4} & 83.4 & 84.3 & 91.2 & 80.5\\
	\specialrule{.1em}{.05em}{.05em}
	\end{tabular}
	\vspace*{-2mm}
\end{table}

\begin{table}[t]
	\centering
	\setlength\tabcolsep{1.0pt}
	\def\arraystretch{1.0}
	\caption{Sequence-wise PCK$_{\rm abs}$ on the MuPoTS-3D dataset.}
	\label{table:mupots_abs20}
	\begin{tabular}{L{2.5cm}*{11}{C{0.73cm}}}
	\multicolumn{12}{l}{\textit{Accuracy for all groundtruths}} \\
	\specialrule{.1em}{.05em}{.05em}
     & S1 & S2 & S3 & S4 & S5 & S6 & S7 & S8 & S9 & S10 \\ \hline
    Moon et al.~\cite{Moon_2019_ICCV_3DMPPE} & \textbf{59.5} & \textbf{44.7} & \textbf{51.4} & \textbf{46.0} & \textbf{52.2} & \textbf{27.4} & 23.7 & \textbf{26.4} & \textbf{39.1} & 23.6 & \\
    \textbf{Ours} & 41.6 & 33.4 & 45.6 & 16.2 & 48.8 & 25.8 & \textbf{46.5} & 13.4 & 36.7 & \textbf{73.5} &\\
    \specialrule{.1em}{.05em}{.05em}
    & S11 & S12 & S13 & S14 & S15 & S16 & S17 & S18 & S19 & S20 & Avg. \\ \hline
    Moon et al.~\cite{Moon_2019_ICCV_3DMPPE} & 18.3 & 14.9 & \textbf{38.2} & 26.5 & 36.8 & 23.4 & 14.4 & \textbf{19.7} & 18.8 & 25.1 & 31.5 \\
	\textbf{Ours} & \textbf{43.6} & \textbf{22.7} & 21.9 & \textbf{26.7} & \textbf{47.1} & \textbf{32.5} & \textbf{31.4} & 18.0 & \textbf{33.8} & \textbf{47.8} & \textbf{35.4}\\
    \specialrule{.1em}{.05em}{.05em}
    \\
    \multicolumn{12}{l}{\textit{Accuracy for matched groundtruths}} \\
    \specialrule{.1em}{.05em}{.05em}
     & S1  & S2 & S3 & S4 & S5 & S6 & S7 & S8 & S9 & S10 \\ \hline
	Moon et al.~\cite{Moon_2019_ICCV_3DMPPE} & \textbf{59.5} & \textbf{45.3} & \textbf{51.4} & \textbf{46.2} & 53.0 & \textbf{27.4} & 23.7 & \textbf{26.4} & \textbf{39.1} & 23.6 & \\
    \textbf{Ours} & 42.1 & 41.4 & 46.5 & 16.3 & \textbf{53.1} & 26.4 & \textbf{47.5} & 18.7 & 36.7 & \textbf{73.5} &\\
    \specialrule{.1em}{.05em}{.05em}
    & S11 & S12 & S13 & S14 & S15 & S16 & S17 & S18 & S19 & S20 & Avg. \\ \hline
    Moon et al.~\cite{Moon_2019_ICCV_3DMPPE} & 18.3 & 14.9 & \textbf{38.2} & 29.5 & 36.8 & 23.6 & 14.4 & 20.0 & 18.8 & 25.4 & 31.8 \\
	\textbf{Ours} & \textbf{46.0} & \textbf{22.7} & 24.3 & \textbf{38.9} & \textbf{47.5} & \textbf{34.2} & \textbf{35.0} & \textbf{20.1} & \textbf{38.7} & \textbf{64.8} & \textbf{38.7}\\
	\specialrule{.1em}{.05em}{.05em}
	\end{tabular}
\end{table}

\begin{table}[t]
	\centering
	\setlength\tabcolsep{1.0pt}
	\def\arraystretch{1.0}
	\caption{PCK$_{\rm rel}$ on the MuPoTS-3D dataset for all groundtruths. 
	`T' denotes top-down methods while `B' denotes bottom-up methods.}
	\label{table:mupots_rel_all}
	\begin{tabular}{L{2.5cm}L{0.5cm} *{11}{C{0.73cm}}}
	\specialrule{.1em}{.05em}{.05em}
    & & S1 & S2 & S3 & S4 & S5 & S6 & S7 & S8 & S9 & S10 \\ \hline
    Rogez et al.~\cite{rogez2017lcr} & T & 67.7 & 49.8 & 53.4 & 59.1 & 67.5 & 22.8 & 43.7 & 49.9 & 31.1 & 78.1 & \\
    Rogez et al.~\cite{rogez2019lcr} & T & 87.3 & 61.9 & 67.9 & 74.6 & 78.8 & 48.9 & 58.3 & 59.7 & 78.1 & 89.5 &  \\
    Dabral et al.~\cite{dabral2019multi} & T & 85.1 & 67.9 & 73.5 & 76.2 & 74.9 & 52.5 & 65.7 & 63.6 & 56.3 & 77.8 & \\
	Moon et al.~\cite{Moon_2019_ICCV_3DMPPE} & T & \textbf{94.4} & \textbf{77.5} & \textbf{79.0} & \textbf{81.9} & \textbf{85.3} & \textbf{72.8} & 81.9 & \textbf{75.7} & \textbf{90.2} & \textbf{90.4} & \\
	Mehta et al.~\cite{mehta2018single} & B & 81.0 & 59.9 & 64.4 & 62.8 & 68.0 & 30.3 & 65.0 & 59.2 & 64.1 & 83.9 &  \\
	Mehta et al.~\cite{mehta2019xnect} & B & 88.4 & 65.1 & 68.2 & 72.5 & 76.2 & 46.2 & 65.8 & 64.1 & 75.1 & 82.4 &  \\
    \textbf{Ours} & B & 88.8 & 71.2 & 77.4 & 77.7 & 80.6 & 49.9 & \textbf{86.6} & 51.3 & 70.3 & 89.2  &\\
    \specialrule{.1em}{.05em}{.05em}
    & & S11 & S12 & S13 & S14 & S15 & S16 & S17 & S18 & S19 & S20 & Avg. \\ \hline
    Rogez et al.~\cite{rogez2017lcr} & T & 50.2 & 51.0 & 51.6 & 49.3 & 56.2 & 66.5 & 65.2 & 62.9 & 66.1 & 59.1 & 53.8 \\
    Rogez et al.~\cite{rogez2019lcr} & T & 69.2 & 73.8 & 66.2 & 56.0 & 74.1 & 82.1 & 78.1 & 72.6 & 73.1 & 61.0 & 70.6 \\
    Dabral et al.~\cite{dabral2019multi} & T & 76.4 & 70.1 & 65.3 & 51.7 & 69.5 & 87.0 & 82.1 & 80.3 & 78.5 & 70.7 & 71.3 \\
    Moon et al.~\cite{Moon_2019_ICCV_3DMPPE} & T & \textbf{79.2} & 79.9 & \textbf{75.1} & \textbf{72.7} & \textbf{81.1} & \textbf{89.9} & \textbf{89.6} & \textbf{81.8} & \textbf{81.7} & \textbf{76.2} & \textbf{81.8} \\
    Mehta et al.~\cite{mehta2018single} & B & 67.2 & 68.3 & 60.6 & 56.5 & 69.9 & 79.4 & 79.6 & 66.1 & 66.3 & 63.5 & 65.0 \\
	Mehta et al.~\cite{mehta2019xnect} & B & 74.1 & 72.4 & 64.4 & 58.8 & 73.7 & 80.4 & 84.3 & 67.2 & 74.3 & 67.8 & 70.4 \\
	\textbf{Ours} & B & 72.3 & \textbf{81.7} & 63.6 & 44.8 & 79.7 & 86.9 & 81.0 & 75.2 & 73.6 & 67.2 & 73.5\\
	\specialrule{.1em}{.05em}{.05em}
	\end{tabular}
\end{table}

\begin{table}[t]
	\centering
	\setlength\tabcolsep{1.0pt}
	\def\arraystretch{1.0}
	\caption{MPJPE Results on Human3.6M dataset. Note that there is no groundtruth bounding box information in inference time.}
	\label{table:h36m}
	\begin{tabular}{L{3cm}C{1.7cm}}
	\specialrule{.1em}{.05em}{.05em}
	Method & MPJPE \\ \hline
    Rogez et al. \cite{rogez2017lcr} & 87.7 \\
    Mehta et al. \cite{mehta2018single} & 69.9 \\
    Dabral et al. \cite{dabral2019multi} & 65.2 \\
    Mehta et al. \cite{mehta2019xnect} & 63.6 \\
    Rogez et al. \cite{rogez2019lcr} & 63.5 \\
    Moon et al. \cite{Moon_2019_ICCV_3DMPPE} & 54.4 \\
    \textbf{Ours} & \textbf{54.1} \\
	
	\specialrule{.1em}{.05em}{.05em}
	\end{tabular}
\end{table}

\begin{table}[t]
	\centering
	\setlength\tabcolsep{1.0pt}
	\def\arraystretch{1.0}
	\caption{Ablation study of the structure design on the MuPoTS-3D dataset. Note that our full model consists of root depth, relative depth and 2D branches.}
	\label{table:ablation_root}
	\begin{tabular}{L{5.5cm}|*{5}{C{1.2cm}}}
	\specialrule{.1em}{.05em}{.05em}
    Design & Recall & PCK$_{\rm root}$ & PCK$_{\rm abs}$ & PCK$_{\rm rel}$ & PCOD\\\hline
    Full Model & \textbf{92.3} & \textbf{45.5} & \textbf{38.7} & \textbf{80.5} & \textbf{97.0} \\
    Root Depth Only & 85.0 & 29.9 & - & - & 88.3 \\
	Root Depth + 2D Branches & 92.1 & 43.6 & - & - & 96.7 \\
	
	\specialrule{.1em}{.05em}{.05em}
	\end{tabular}
\end{table}

\clearpage
\bibliographystyle{splncs04}
\bibliography{egbib}
\end{document}